\documentclass[nohyperref]{article}

\usepackage{microtype}
\usepackage{graphicx}
\usepackage{subfigure}
\usepackage{booktabs} 

\usepackage{hyperref}

\usepackage[accepted]{icml2022}

\usepackage{amsmath}
\usepackage{amssymb}
\usepackage{mathtools}
\usepackage{amsthm}

\usepackage[capitalize,noabbrev]{cleveref}

\usepackage{adjustbox} 
\usepackage{booktabs}  
\usepackage{multirow}  
\usepackage{threeparttable}

\usepackage{array}
\newcolumntype{?}{!{\vrule width 1.7pt}}
\usepackage{enumitem}

\theoremstyle{plain}

\theoremstyle{definition}

\theoremstyle{remark}

\usepackage[textsize=tiny]{todonotes}

\icmltitlerunning{Cut Inner Layers: A Structured Pruning Strategy for Efficient U-Net GANs}

\begin{document}

\twocolumn[
\icmltitle{Cut Inner Layers: A Structured Pruning Strategy for Efficient U-Net GANs}




\begin{icmlauthorlist}
\icmlauthor{Bo-Kyeong Kim}{nota}
\icmlauthor{Shinkook Choi}{nota}
\icmlauthor{Hancheol Park}{nota}
\end{icmlauthorlist}

\icmlaffiliation{nota}{Nota Inc., Seoul, South Korea}

\icmlcorrespondingauthor{Bo-Kyeong Kim}{bokyeong1015@gmail.com}

\icmlkeywords{Structured Pruning, U-Net, Generative Adversarial Network, GAN}

\vskip 0.3in
]



\printAffiliationsAndNotice{}  

\begin{abstract}

Pruning effectively compresses overparameterized models. Despite the success of pruning methods for discriminative models, applying them for generative models has been relatively rarely approached. This study conducts structured pruning on U-Net generators of conditional GANs. A per-layer sensitivity analysis confirms that many unnecessary filters exist in the innermost layers near the bottleneck and can be substantially pruned. Based on this observation, we prune these filters from multiple inner layers or suggest alternative architectures by completely eliminating the layers. We evaluate our approach with Pix2Pix for image-to-image translation and Wav2Lip for speech-driven talking face generation. Our method outperforms global pruning baselines, demonstrating the importance of properly considering where to prune for U-Net generators.

\end{abstract}

\section{Introduction} \label{sec:intro}

Developing lightweight neural networks has attracted considerable attention for their efficient deployment in real-world applications. A popular method for network compression is pruning \cite{lecun1989optimal, wen2016learning}, which removes unimportant weights from overparameterized models. Numerous studies have successfully applied pruning methods for discriminative models (e.g., image classifiers \cite{frankle2018lottery, ning2020dsa, yeom2021pruning} and object detectors \cite{shih2019real, ghosh2019deep, xie2020localization}); however, applying them for generative models has been relatively less investigated.

\begin{figure}[t]
  \centering
   \includegraphics[width=0.9\columnwidth]{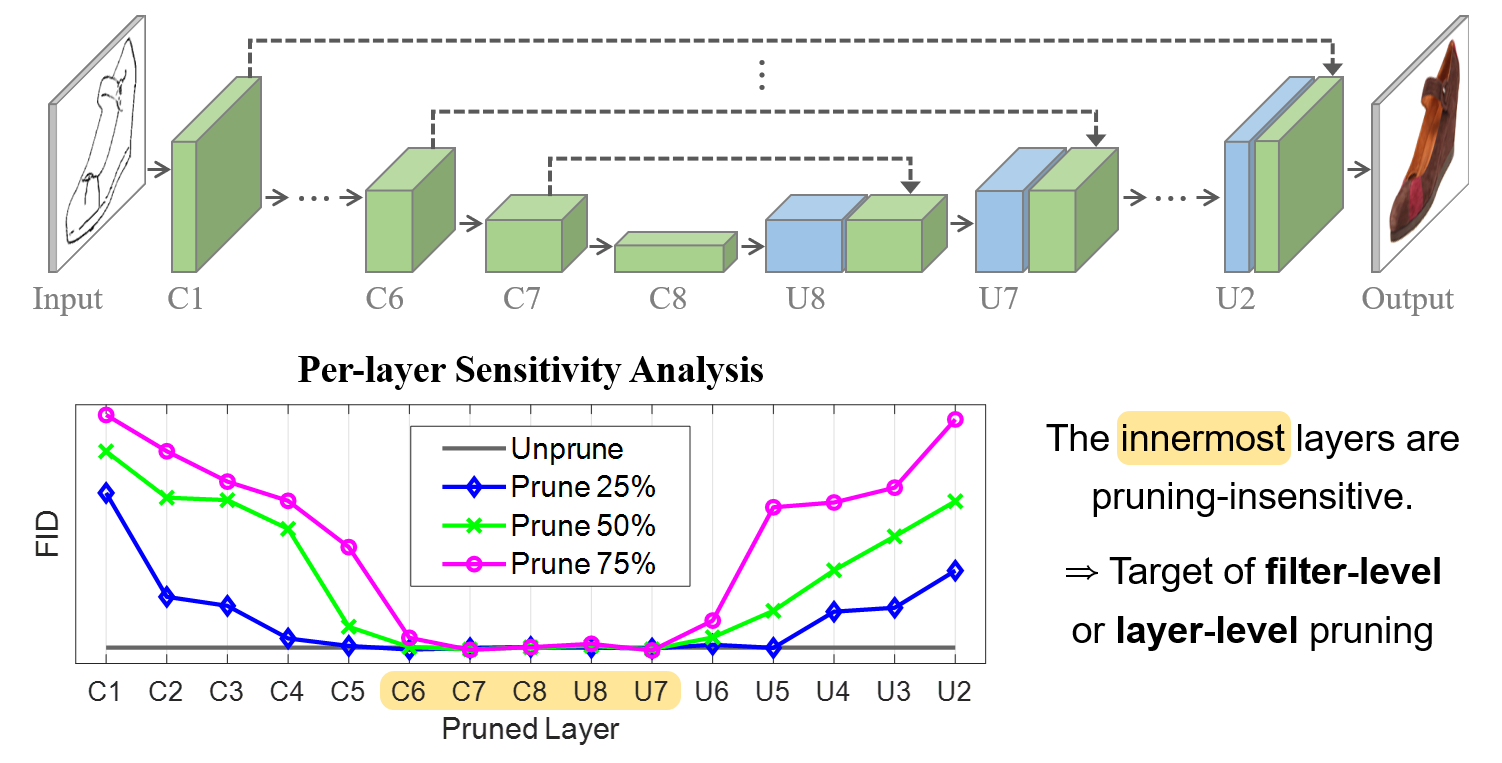}
   \caption{Illustration of (top) a U-Net generator and (bottom) a per-layer pruning sensitivity analysis. A lower FID indicates a better generation quality. The analysis reveals that the innermost layers in U-Net generators have many unimportant filters, which can be pruned without noticeable performance degradation.}
   \label{fig_overview}
\end{figure}

\begin{figure*}[t]
\centering
\begin{center}
     \includegraphics[width=2\columnwidth]{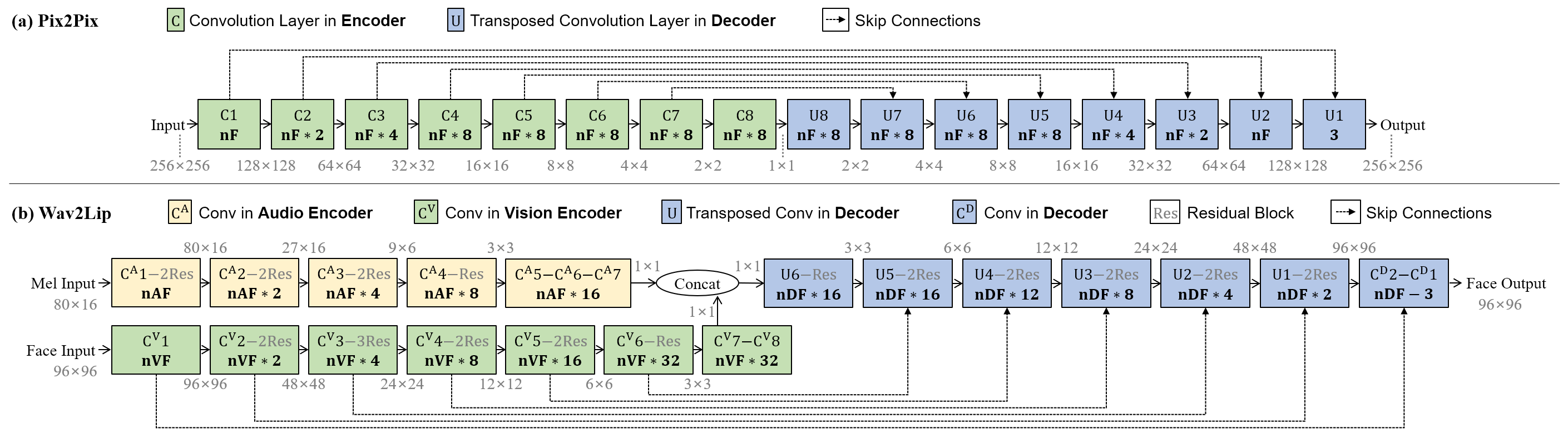}
    \caption{Generator architecture of (a) Pix2Pix and (b) Wav2Lip. Each layer is denoted by the type of convolution and an identification number (e.g., $\textrm{C2}$ denotes the second convolutional layer). The number of base filters for Pix2Pix is denoted by $\mathrm{nF}=64$, and that for Wav2Lip is denoted by $(\mathrm{nVF}, \mathrm{nAF}, \mathrm{nDF})=(16, 32, 32)$. Skip connections between the encoder and the decoder are indicated with dotted lines. For brevity, we omit the normalization layers and nonlinear functions.}

    \label{fig_network}
\end{center}
\end{figure*}

\begin{figure*}[t]
\centering
\begin{center}
     \includegraphics[width=2\columnwidth]{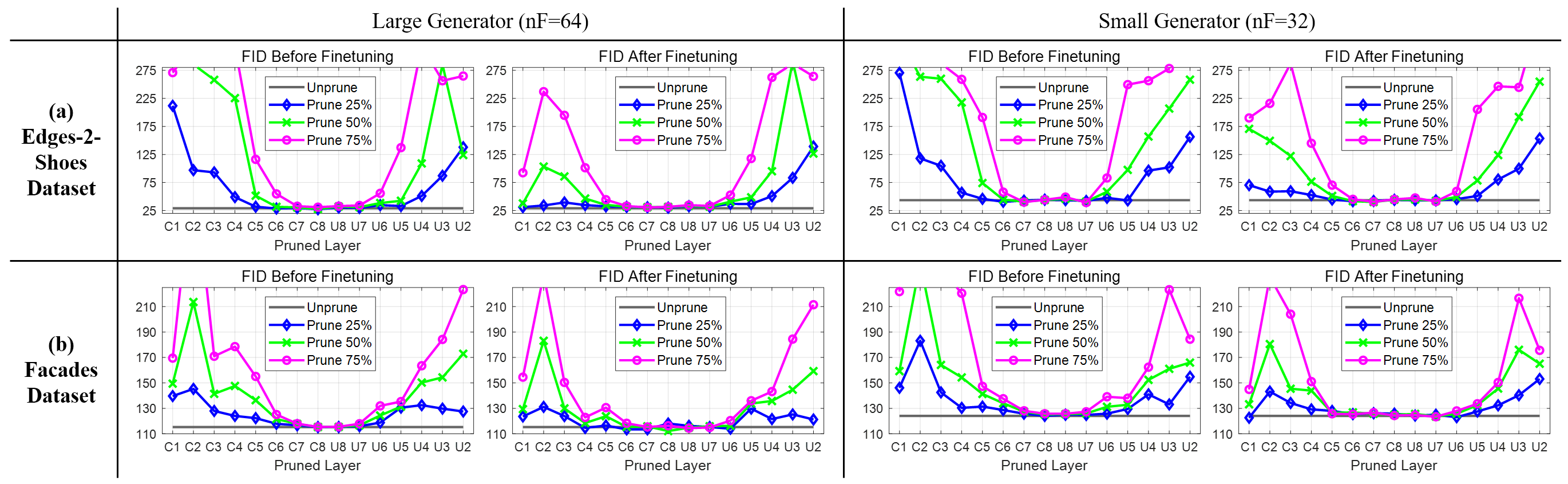}
    \caption{Layer-wise sensitivity analysis on the (a) Edges2Shoes (edges→photo) and (b) Facades (architectural labels→photo) datasets. A lower FID indicates a better generation quality.}
    \label{fig_per_layer_sensitivity}
\end{center}
\end{figure*}

In this study, we focus on pruning of U-Net \cite{ronneberger2015u} generators that have been widely used in generative adversarial networks (GANs) \cite{goodfellow2014generative} for conditional image synthesis tasks \cite{yang2020towards, pix2pix2017, prajwal2020lip}. As shown in \Cref{fig_overview}, a U-Net consists of an encoder and a decoder, where skip connections shuttle context information across the network. It contains a huge number of parameters, all of which may not be necessary for high-fidelity image generation. Through pruning, we not only compress the generator but also deepen our understanding on the network behavior.

A layer-wise sensitivity analysis for understanding how pruning of each layer affects the performance has been conducted merely on classification models \cite{ning2020dsa, li2016pruning, lin2019toward}. Applying such an analysis for GANs is arguably more challenging because of the difficulty in retraining and evaluating pruned generators. To the best of our knowledge, this is the first attempt to analyze the pruning sensitivity of GAN generators. Our analysis reveals that many filters in the innermost layers of the U-Net are redundant and can be removed. We observe this phenomenon consistently with different datasets, network capacities, pruning criteria, and evaluation metrics.

Based on the analysis, we perform filter pruning of multiple innermost layers or even remove these layers entirely. We evaluate our approach for the U-Net generator of Pix2Pix \cite{pix2pix2017} on the Edges2Shoes \cite{data_e2s} and Facades \cite{data_fcd} datasets and for that of Wav2Lip \cite{prajwal2020lip} on the LRS3 dataset \cite{data_lrs3}. Our method outperforms global pruning baselines that are commonly used for classification models, indicating that properly considering where to prune for the U-Net is important.

\section{Related Work} \label{sec:rel_work}

Although there have been several studies for learning efficient GANs (e.g., using neural architecture search \cite{gao2020adversarialnas, gong2019autogan, fu2020autogan, hou2020slimmable} or network quantization \cite{wang2020gan, wang2019qgan, wan2020deep, andreev2021quantization}), we mainly discuss pruning-based approaches that are often combined with knowledge distillation (KD) \cite{hinton2015distilling, romero2014fitnets} to improve the performance.

In the KD setups of \citet{li2022learning, li2021revisiting, jin2021teachers, liu2021content, li2020gan}, pruned generators that act as students are trained under the guidance of original large generators that act as teachers. To derive pruned models, they inject sparsity constraints during training the original generator \cite{li2022learning, li2021revisiting} or redesign the original model to provide an architectural search space \cite{jin2021teachers}; however, these methods become difficult or impossible to apply when the compression of already pretrained generators is needed. We consider a practical case for pruning of pretrained GANs obtained without sparsity regularization or redesigning steps.

Instead of using pruned generators as the initialization of students \cite{li2021revisiting, jin2021teachers, liu2021content}, some studies \cite{wang2020gan, shu2019co} infer compressed student architectures during the KD process. Unstructured pruning has also been applied to compress GANs with the lottery ticket hypothesis \cite{chen2021gans, chen2021data} or with a KD strategy \cite{yu2020self}.


\begin{table*}[t]
\centering
\begin{adjustbox}{max width=2\columnwidth}

\begin{tabular}{cccc|ccc|ccc}

\specialrule{.2em}{.1em}{.1em}

\multicolumn{4}{c|}{Method}                                                                                                                                                                                                                                                       & \multicolumn{3}{c|}{Large Generator (nF=64)}                                               & \multicolumn{3}{c}{Small Generator (nF=32)}                                             \\

\hline

\multicolumn{2}{c|}{Category}                                                                                                                                                                   & \multicolumn{1}{c|}{Bottleneck}   & Pruned Layer/Ratio                          & \multicolumn{1}{c|}{FID (↓)}       & \multicolumn{1}{c|}{\#Params}       & MACs            & \multicolumn{1}{c|}{FID (↓)}       & \multicolumn{1}{c|}{\#Params}      & MACs           \\

\specialrule{.2em}{.1em}{.1em}

\multicolumn{2}{c|}{Original Pix2Pix}                                                                                                                                                           & \multicolumn{1}{c|}{1×1}          & -                                           & \multicolumn{1}{c|}{29.0}          & \multicolumn{1}{c|}{54.4M}          & 18.14G          & \multicolumn{1}{c|}{43.5}          & \multicolumn{1}{c|}{13.6M}         & 4.65G          \\

\specialrule{.2em}{.1em}{.1em}

\multicolumn{1}{c|}{\multirow{6}{*}{\begin{tabular}[c]{@{}c@{}}Global\\ Pruning\end{tabular}}} & \multicolumn{1}{c|}{\multirow{2}{*}{Uniform}}                                                  & \multicolumn{1}{c|}{1×1}          & All Layers/1\%                              & \multicolumn{1}{c|}{33.5}          & \multicolumn{1}{c|}{53.4M}          & 17.88G          & \multicolumn{1}{c|}{43.9}          & \multicolumn{1}{c|}{13.4M}         & 4.61G          \\ \cline{3-10} 
\multicolumn{1}{c|}{}                                                                          & \multicolumn{1}{c|}{}                                                                          & \multicolumn{1}{c|}{1×1}          & All Layers/2\%                              & \multicolumn{1}{c|}{59.3}          & \multicolumn{1}{c|}{52.3M}          & 17.51G          & \multicolumn{1}{c|}{46.6}          & \multicolumn{1}{c|}{13.1M}         & 4.52G          \\ 

\cline{2-10} 
\noalign{\vskip\doublerulesep
         \vskip-\arrayrulewidth}
\cline{2-10}

\multicolumn{1}{c|}{}                                                                          & \multicolumn{1}{c|}   
{\multirow{2}{*}{\begin{tabular}[c]{@{}c@{}}Uniform+\\\cite{gale2019state}\end{tabular}}}                                                & \multicolumn{1}{c|}{1×1}          & All Layers Except C1 \& U2/2\%              & \multicolumn{1}{c|}{38.7}          & \multicolumn{1}{c|}{52.3M}          & 17.59G          & \multicolumn{1}{c|}{46.6}         & \multicolumn{1}{c|}{13.1M}         & 4.52G          \\ \cline{3-10} 
\multicolumn{1}{c|}{}                                                                          & \multicolumn{1}{c|}{}                                                                          & \multicolumn{1}{c|}{1×1}          & All Layers Except C1 \& U2/3\%              & \multicolumn{1}{c|}{41.1}          & \multicolumn{1}{c|}{51.3M}          & 17.33G          & \multicolumn{1}{c|}{48.0}         & \multicolumn{1}{c|}{12.9M}         & 4.48G          \\ 

\cline{2-10} 
\noalign{\vskip\doublerulesep
         \vskip-\arrayrulewidth}
\cline{2-10}

\multicolumn{1}{c|}{}                                                                          & \multicolumn{1}{c|}{\multirow{2}{*}{\begin{tabular}[c]{@{}c@{}}Structured LAMP\\\cite{lee2020layer}
\end{tabular}}}  & \multicolumn{1}{c|}{1×1}          & LAMP-based Selection/1\%                    & \multicolumn{1}{c|}{47.0}          & \multicolumn{1}{c|}{53.4M}          & 16.78G          & \multicolumn{1}{c|}{66.8}          & \multicolumn{1}{c|}{13.4M}         & 4.35G          \\ \cline{3-10} 
\multicolumn{1}{c|}{}                                                                          & \multicolumn{1}{c|}{}                                                                          & \multicolumn{1}{c|}{1×1}          & LAMP-based Selection/2\%                    & \multicolumn{1}{c|}{75.9}          & \multicolumn{1}{c|}{52.4M}          & 15.39G          & \multicolumn{1}{c|}{118.3}         & \multicolumn{1}{c|}{13.1M}         & 4.02G          \\

\specialrule{.2em}{.1em}{.1em}

\multicolumn{1}{c|}{\multirow{4}{*}{Ours}}                                                     & \multicolumn{1}{c|}{\multirow{2}{*}{\begin{tabular}[c]{@{}c@{}}Filter\\ Pruning\end{tabular}}} & \multicolumn{1}{c|}{1×1}          & \{C6-C7-C8\}/50\%                           & \multicolumn{1}{c|}{30.4}          & \multicolumn{1}{c|}{39.7M}          & 17.92G          & \multicolumn{1}{c|}{43.7}          & \multicolumn{1}{c|}{9.9M}          & 4.59G          \\ \cline{3-10} 
\multicolumn{1}{c|}{}                                                                          & \multicolumn{1}{c|}{}                                                                          & \multicolumn{1}{c|}{\textbf{1×1}} & \textbf{\{C6-C7-C8\}/50\% + \{U8-U7\}/25\%} & \multicolumn{1}{c|}{\textbf{31.9}} & \multicolumn{1}{c|}{\textbf{35.8M}} & \textbf{17.81G} & \multicolumn{1}{c|}{\textbf{44.4}} & \multicolumn{1}{c|}{\textbf{9.0M}} & \textbf{4.57G} \\ 

\cline{2-10} 
\noalign{\vskip\doublerulesep
         \vskip-\arrayrulewidth}
\cline{2-10} 

\multicolumn{1}{c|}{}                                                                          & \multicolumn{1}{c|}{\multirow{2}{*}{\begin{tabular}[c]{@{}c@{}}Layer\\ Removal\end{tabular}}}  & \multicolumn{1}{c|}{2×2}          & \{C8-U8\}/100\%                             & \multicolumn{1}{c|}{28.3}          & \multicolumn{1}{c|}{41.8M}          & 18.06G          & \multicolumn{1}{c|}{39.9}          & \multicolumn{1}{c|}{10.5M}         & 4.63G          \\ \cline{3-10} 
\multicolumn{1}{c|}{}                                                                          & \multicolumn{1}{c|}{}                                                                          & \multicolumn{1}{c|}{\textbf{4×4}} & \textbf{\{C7-C8-U8-U7\}/100\%}              & \multicolumn{1}{c|}{\textbf{29.7}} & \multicolumn{1}{c|}{\textbf{29.2M}} & \textbf{17.70G} & \multicolumn{1}{c|}{\textbf{41.9}} & \multicolumn{1}{c|}{\textbf{7.3M}} & \textbf{4.54G} \\

\specialrule{.2em}{.1em}{.1em}

\end{tabular}

\end{adjustbox}

\caption{Quantitative results on the Edges2Shoes (edges→photo) dataset. The models marked in bold face provide a good performance-efficiency trade-off (see \Cref{fig_visualEx_pix2pix}(a) for their visual results).}
\label{table:e2s_result}
\end{table*}


\begin{table*}[t]
\centering
\begin{adjustbox}{max width=2\columnwidth}

\begin{tabular}{cccc|ccc|ccc}

\specialrule{.2em}{.1em}{.1em}

\multicolumn{4}{c|}{Method}                                                                                                                                                                                                                                                             & \multicolumn{3}{c|}{Large Generator (nF=64)}                                                & \multicolumn{3}{c}{Small Generator (nF=32)}                                              \\ \hline
\multicolumn{2}{c|}{Category}                                                                                                                                                                   & \multicolumn{1}{c|}{Bottleneck}   & Pruned Layer/Ratio                                & \multicolumn{1}{c|}{FID (↓)}        & \multicolumn{1}{c|}{\#Params}       & MACs            & \multicolumn{1}{c|}{FID (↓)}        & \multicolumn{1}{c|}{\#Params}      & MACs           \\

\specialrule{.2em}{.1em}{.1em}

\multicolumn{2}{c|}{Original Pix2Pix}                                                                                                                                                           & \multicolumn{1}{c|}{1×1}          & -                                                 & \multicolumn{1}{c|}{115.3}          & \multicolumn{1}{c|}{54.4M}          & 18.14G          & \multicolumn{1}{c|}{124.0}          & \multicolumn{1}{c|}{13.6M}         & 4.65G          \\

\specialrule{.2em}{.1em}{.1em}

\multicolumn{1}{c|}{\multirow{6}{*}{\begin{tabular}[c]{@{}c@{}}Global\\ Pruning\end{tabular}}} & \multicolumn{1}{c|}{\multirow{2}{*}{Uniform}}                                                  & \multicolumn{1}{c|}{1×1}          & All Layers/2\%                                    & \multicolumn{1}{c|}{118.1}          & \multicolumn{1}{c|}{52.3M}          & 17.51G          & \multicolumn{1}{c|}{128.4}          & \multicolumn{1}{c|}{13.1M}         & 4.52G          \\ \cline{3-10} 
\multicolumn{1}{c|}{}                                                                          & \multicolumn{1}{c|}{}                                                                          & \multicolumn{1}{c|}{1×1}          & All Layers/4\%                                    & \multicolumn{1}{c|}{127.9}          & \multicolumn{1}{c|}{50.2M}          & 16.81G          & \multicolumn{1}{c|}{139.1}          & \multicolumn{1}{c|}{12.6M}         & 4.33G          \\ 

\cline{2-10} 
\noalign{\vskip\doublerulesep
         \vskip-\arrayrulewidth}
\cline{2-10}

\multicolumn{1}{c|}{}                                                                          & \multicolumn{1}{c|}   
{\multirow{2}{*}{\begin{tabular}[c]{@{}c@{}}Uniform+\\\cite{gale2019state}\end{tabular}}}

& \multicolumn{1}{c|}{1×1}          & All Layers Except C1 \& U2/2\%                    & \multicolumn{1}{c|}{119.4}          & \multicolumn{1}{c|}{52.3M}          & 17.59G          & \multicolumn{1}{c|}{128.4}         & \multicolumn{1}{c|}{13.1M}         & 4.52G          \\ \cline{3-10} 
\multicolumn{1}{c|}{}                                                                          & \multicolumn{1}{c|}{}                                                                          & \multicolumn{1}{c|}{1×1}          & All Layers Except C1 \& U2/4\%                    & \multicolumn{1}{c|}{123.4}          & \multicolumn{1}{c|}{50.3M}          & 16.97G          & \multicolumn{1}{c|}{135.3}          & \multicolumn{1}{c|}{12.6M}         & 4.37G          \\ 

\cline{2-10} 
\noalign{\vskip\doublerulesep
         \vskip-\arrayrulewidth}
\cline{2-10}

\multicolumn{1}{c|}{}                                                                          & \multicolumn{1}{c|}{\multirow{2}{*}{\begin{tabular}[c]{@{}c@{}}Structured LAMP\\\cite{lee2020layer}\end{tabular}}}  & \multicolumn{1}{c|}{1×1}          & LAMP-based Selection/4\%                          & \multicolumn{1}{c|}{115.2}          & \multicolumn{1}{c|}{50.6M}          & 18.11G          & \multicolumn{1}{c|}{127.1}          & \multicolumn{1}{c|}{12.6M}         & 4.57G          \\ \cline{3-10} 
\multicolumn{1}{c|}{}                                                                          & \multicolumn{1}{c|}{}                                                                          & \multicolumn{1}{c|}{1×1}          & LAMP-based Selection/6\%                          & \multicolumn{1}{c|}{118.9}          & \multicolumn{1}{c|}{48.2M}          & 17.75G          & \multicolumn{1}{c|}{127.6}          & \multicolumn{1}{c|}{12.0M}         & 4.49G          \\

\specialrule{.2em}{.1em}{.1em} 

\multicolumn{1}{c|}{\multirow{4}{*}{Ours}}                                                     & \multicolumn{1}{c|}{\multirow{2}{*}{\begin{tabular}[c]{@{}c@{}}Filter\\ Pruning\end{tabular}}} & \multicolumn{1}{c|}{1×1}          & C6/50\% + \{C7-C8-U8-U7\}/75\%                    & \multicolumn{1}{c|}{113.2}          & \multicolumn{1}{c|}{27.7M}          & 17.61G          & \multicolumn{1}{c|}{124.7}          & \multicolumn{1}{c|}{6.9M}          & 4.52G          \\ \cline{3-10} 
\multicolumn{1}{c|}{}                                                                          & \multicolumn{1}{c|}{}                                                                          & \multicolumn{1}{c|}{\textbf{1×1}} & \textbf{C6/50\% + \{C7-C8-U8-U7\}/75\% + U6/25\%} & \multicolumn{1}{c|}{\textbf{114.0}} & \multicolumn{1}{c|}{\textbf{25.8M}} & \textbf{17.30G} & \multicolumn{1}{c|}{\textbf{125.3}} & \multicolumn{1}{c|}{\textbf{6.5M}} & \textbf{4.44G} \\ 

\cline{2-10} 
\noalign{\vskip\doublerulesep
         \vskip-\arrayrulewidth}
\cline{2-10} 

\multicolumn{1}{c|}{}                                                                          & \multicolumn{1}{c|}{\multirow{2}{*}{\begin{tabular}[c]{@{}c@{}}Layer\\ Removal\end{tabular}}}  & \multicolumn{1}{c|}{2×2}          & \{C8-U8\}/100\%                                   & \multicolumn{1}{c|}{107.9}          & \multicolumn{1}{c|}{41.8M}          & 18.06G          & \multicolumn{1}{c|}{108.5}          & \multicolumn{1}{c|}{10.5M}         & 4.63G          \\ \cline{3-10} 
\multicolumn{1}{c|}{}                                                                          & \multicolumn{1}{c|}{}                                                                          & \multicolumn{1}{c|}{\textbf{4×4}} & \textbf{\{C7-C8-U8-U7\}/100\%}                    & \multicolumn{1}{c|}{\textbf{108.7}} & \multicolumn{1}{c|}{\textbf{29.2M}} & \textbf{17.70G} & \multicolumn{1}{c|}{\textbf{108.9}} & \multicolumn{1}{c|}{\textbf{7.3M}} & \textbf{4.54G} \\

\specialrule{.2em}{.1em}{.1em} 

\end{tabular}

\end{adjustbox}

\caption{Quantitative results on the Facades (architectural labels→photo) dataset. The models marked in bold face provide a good performance-efficiency trade-off (see \Cref{fig_visualEx_pix2pix}(b) for their visual results).}

\label{table:facades_result}
\end{table*}


\begin{table*}[t]
\centering

\begin{adjustbox}{max width=2\columnwidth}
\begin{threeparttable}

\begin{tabular}{cc|ccc|cc|cc}

\specialrule{.2em}{.1em}{.1em} 

\multicolumn{2}{c|}{\multirow{2}{*}{Method}}                                                                                & \multicolumn{3}{c|}{Performance}                                                        & \multicolumn{2}{c|}{Computation}                                   & \multicolumn{2}{c}{Latency  (ms / 16 frames)}                   \\ \cline{3-9} 
\multicolumn{2}{c|}{}                                                                                                       & \multicolumn{1}{c|}{FID (↓)}       & \multicolumn{1}{c|}{LSE-D (↓)}     & LSE-C (↑)     & \multicolumn{1}{c|}{\# Params}                  & MACs             & \multicolumn{1}{c|}{CPU}                   & GPU                 \\ 

\specialrule{.2em}{.1em}{.1em} 

\multicolumn{1}{c|}{\multirow{2}{*}{Wav2Lip}}                                                  & Original                   & \multicolumn{1}{c|}{6.16}          & \multicolumn{1}{c|}{6.50}          & 7.44          & \multicolumn{1}{c|}{36.3M (1.0×)}          & 6.21G (1.0×)           & \multicolumn{1}{c|}{1111.5 (1.0×)}         & 24.8 (1.0×)         \\ \cline{2-9} 
\multicolumn{1}{c|}{}                                                                          & Compressed Generator        & \multicolumn{1}{c|}{6.15}          & \multicolumn{1}{c|}{6.92}          & 7.17          & \multicolumn{1}{c|}{5.0M (7.2×)}          & 0.84G (7.4×)          & \multicolumn{1}{c|}{200.2 (5.6×)}          & 7.4 (3.4×)          \\ 

\specialrule{.2em}{.1em}{.1em} 

\multicolumn{1}{c|}{\multirow{3}{*}{\begin{tabular}[c]{@{}c@{}}Global\\ Pruning\end{tabular}}} & Uniform (9\%) $^{\textrm{a}}$     & \multicolumn{1}{c|}{11.62}         & \multicolumn{1}{c|}{7.62}          & 6.19          & \multicolumn{1}{c|}{4.2M (8.7×)}          & 0.72G (8.7×)           & \multicolumn{1}{c|}{170.6 (6.5×) }          &  6.9 (3.6×)        \\ \cline{2-9} 
\multicolumn{1}{c|}{}                                                                          & Uniform (38\%) $^{\textrm{b}}$     & \multicolumn{1}{c|}{28.31}         & \multicolumn{1}{c|}{7.82}          & 5.74          & \multicolumn{1}{c|}{1.9M (18.7×)}         & 0.33G (18.7×)          & \multicolumn{1}{c|}{107.3 (10.4×)}         & 4.3 (5.7×)          \\ \cline{2-9} 
\multicolumn{1}{c|}{}                                                                          & Structured LAMP \cite{lee2020layer} (36\%) $^{\textrm{a,b}}$       & \multicolumn{1}{c|}{8.23}          & \multicolumn{1}{c|}{7.53}          & 6.33          & \multicolumn{1}{c|}{2.0M (18.3×)  }          & 0.71G (8.7×)         & \multicolumn{1}{c|}{167.6 (6.6×)}          & 6.3 (3.9×)          \\ 

\specialrule{.2em}{.1em}{.1em} 

\multicolumn{1}{c|}{Ours}                                                                      & Filter Pruning $^{\textrm{c}}$ & \multicolumn{1}{c|}{\textbf{6.09}} & \multicolumn{1}{c|}{\textbf{7.29}} & \textbf{6.61} & \multicolumn{1}{c|}{\textbf{1.9M (18.9×)}} & \textbf{0.70G (8.9×)} & \multicolumn{1}{c|}{\textbf{164.3 (6.8×)}} & \textbf{6.2 (4.0×)} \\ 

\specialrule{.2em}{.1em}{.1em} 

\end{tabular}

\begin{tablenotes}  
\item[a] Similar MACs with ours.
\item[b] Similar number of parameters with ours.
\item[c] Filter pruning of the innermost layers: 50\% filters of \{$\mathrm{C^A}$5-6-7, $\mathrm{C^V}$6-7-8, $\mathrm{U}$6-5\} and 67\% filters of $\mathrm{U4}$ are removed.
\end{tablenotes}

\end{threeparttable}
\end{adjustbox}

\caption{Quantitative results on the LRS3 dataset. The latency was measured on Xeon Silver 4114 CPU or RTX 2080 Ti GPU. ↑ and ↓ indicate that higher and lower values are better, respectively. The pruning ratio used in global pruning is provided in parentheses.}
\label{table:wav2lip_result}

\end{table*}


\section{Method} \label{sec:method}

\subsection{Magnitude-based Pruning Criterion} 

We primarily use a criterion based on weight magnitude for filter pruning. $\mathcal{F}\in\mathbb{R}^{n_{in} \times n_{out} \times k \times k}$ represents the kernel matrix of a convolutional layer, where $n_{in}$, $n_{out}$, and $k$ denote the number of input and output channels and the spatial kernel size, respectively. For the $i$-th filter, $\mathcal{F}_{i}\in\mathbb{R}^{n_{in} \times k \times k}$, its $L$2-norm magnitude $\| \mathcal{F}_{i} \|_{2}$ is used as an importance score. Because filters with smaller weight magnitudes cause low activation feature maps, removing these filters tends to exhibit little or no performance degradation. In \Cref{sect_gm}, we also investigate another criterion based on the redundancy in convolutional filters \cite{he2019filter}.

\subsection{Layer-wise Sensitivity Analysis} 

To identify which layers contain unnecessary filters in the U-Net generator, we prune the filters of each layer independently and observe the generation performance of the pruned network. We vary the pruning ratio of \{25\%, 50\%, 75\%\} for each layer and repeat this process over all parameterized layers. Such an analysis has been conducted for discriminative models \cite{ning2020dsa, li2016pruning, lin2019toward}; to the best of our knowledge, this is the first layer-wise sensitivity analysis for generative models.

\subsection{Filter-level Pruning}

The sensitivity analysis confirms that the innermost layers of the U-Net generator have a large number of unimportant filters and are insensitive to pruning. Based on this observation, we prune these filters simultaneously from multiple innermost layers to boost computational efficiency. We determine the layers being pruned and the pruning ratio through a grid search over the innermost layers.

\subsection{Layer-level Pruning (Layer Removal)}

We also investigate the effect of layer pruning: we remove the innermost layers entirely and make the redesigned network have wider bottleneck dimensions (e.g., 2×2 or 4×4) instead of 1×1. Based on the symmetric architecture of the U-Net, when eliminating the layers of the encoder, we correspondingly remove the mirrored layers of the decoder (e.g., C8 and U8 in \Cref{fig_network}(a) are removed together).

\subsection{Retraining of Pruned Generators}

Given a pretrained GAN, we run a single pruning-retraining cycle. Retraining a pruned generator is arguably more tricky than retraining a typical classifier, because of taking the discriminator into account. We finetune the pruned generator
while jointly training the pretrained discriminator. In \Cref{appendix_sect_retrain}, we discuss other retraining settings and present their results.

\begin{figure}[t]
\centering
\begin{center}
     \includegraphics[width=\columnwidth]{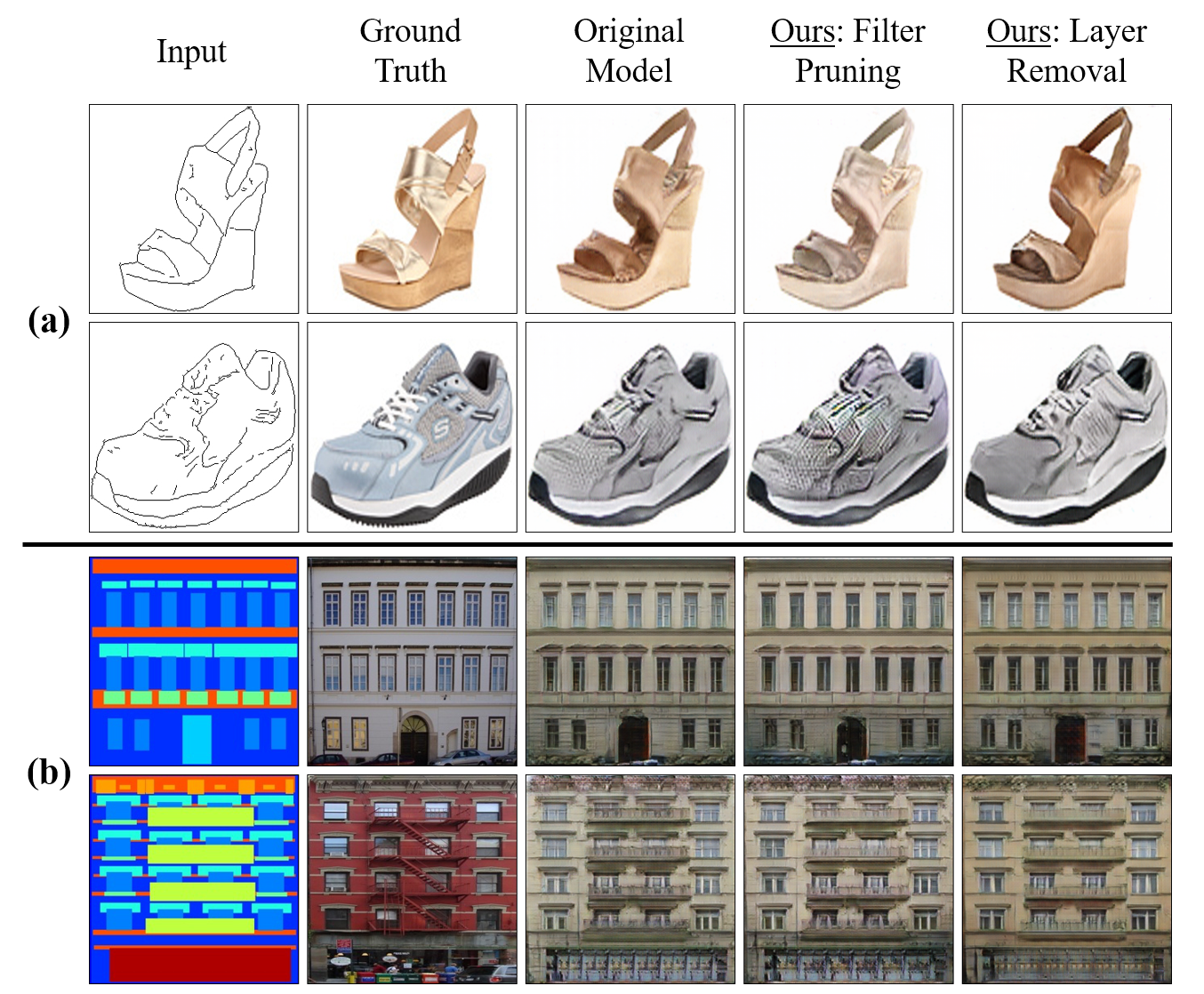}
    \caption{Visual results of Pix2Pix on the (a) Edges2Shoes and (b) Facades datasets.}
    \label{fig_visualEx_pix2pix}
\end{center}
\end{figure} 

\begin{figure}[t]
\centering
\begin{center}
     \includegraphics[width=\columnwidth]{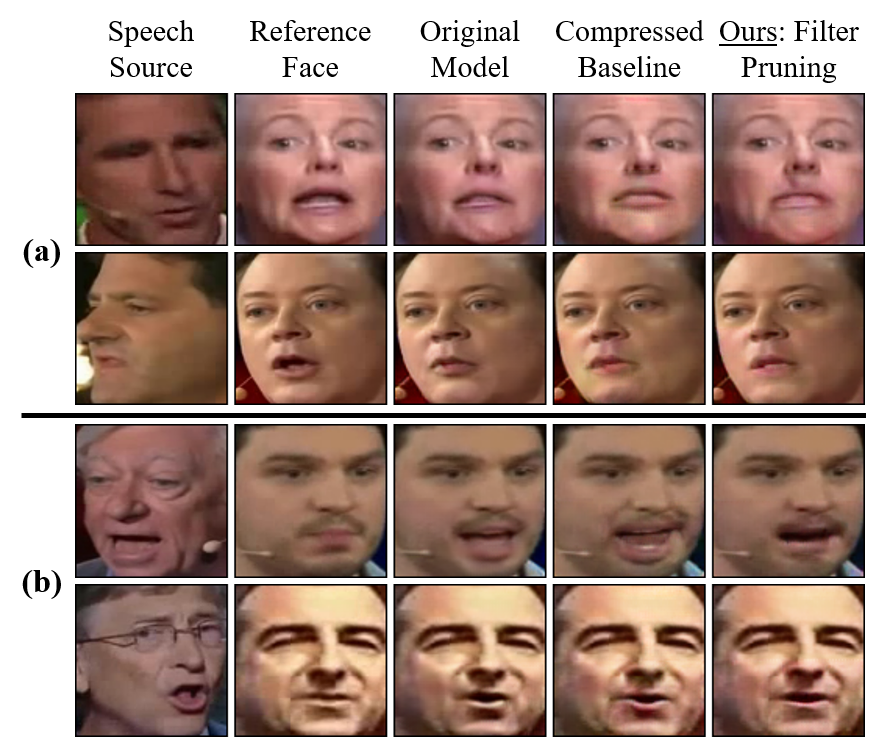}
    \caption{Visual results of Wav2Lip. Following the given speech, the mouth shape of reference faces changes: (a) closed- and (b) open-lip generation.}
    \label{fig_visualEx_wav2lip}
\end{center}
\end{figure} 

\begin{figure*}[h]
\centering
\begin{center}
     \includegraphics[width=2\columnwidth]{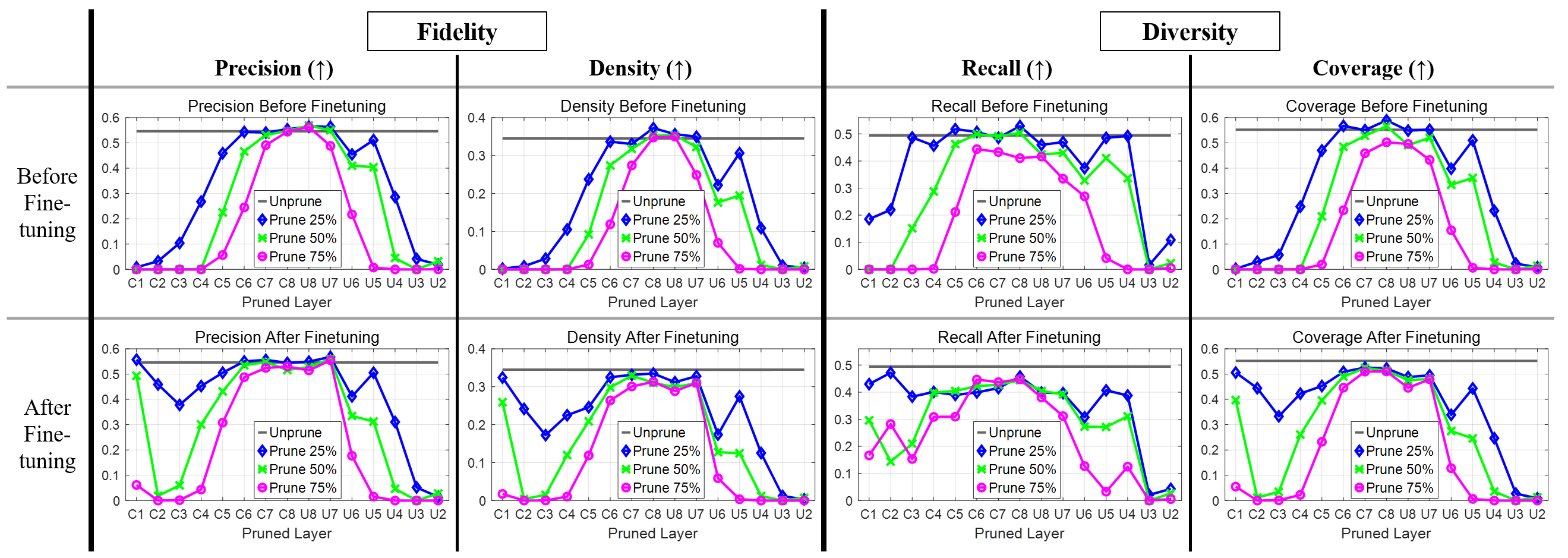}
    \caption{Pruning sensitivity analysis in terms of the fidelity and diversity of generated images. The precision and recall \cite{kynkaanniemi2019improved} and the density and coverage \cite{naeem2020reliable} are used as evaluation metrics. Higher values are better.}
    \label{fig_eval_metrics}
\end{center}
\end{figure*} 

\begin{figure}[t]
\centering
\begin{center}
     \includegraphics[width=\columnwidth]{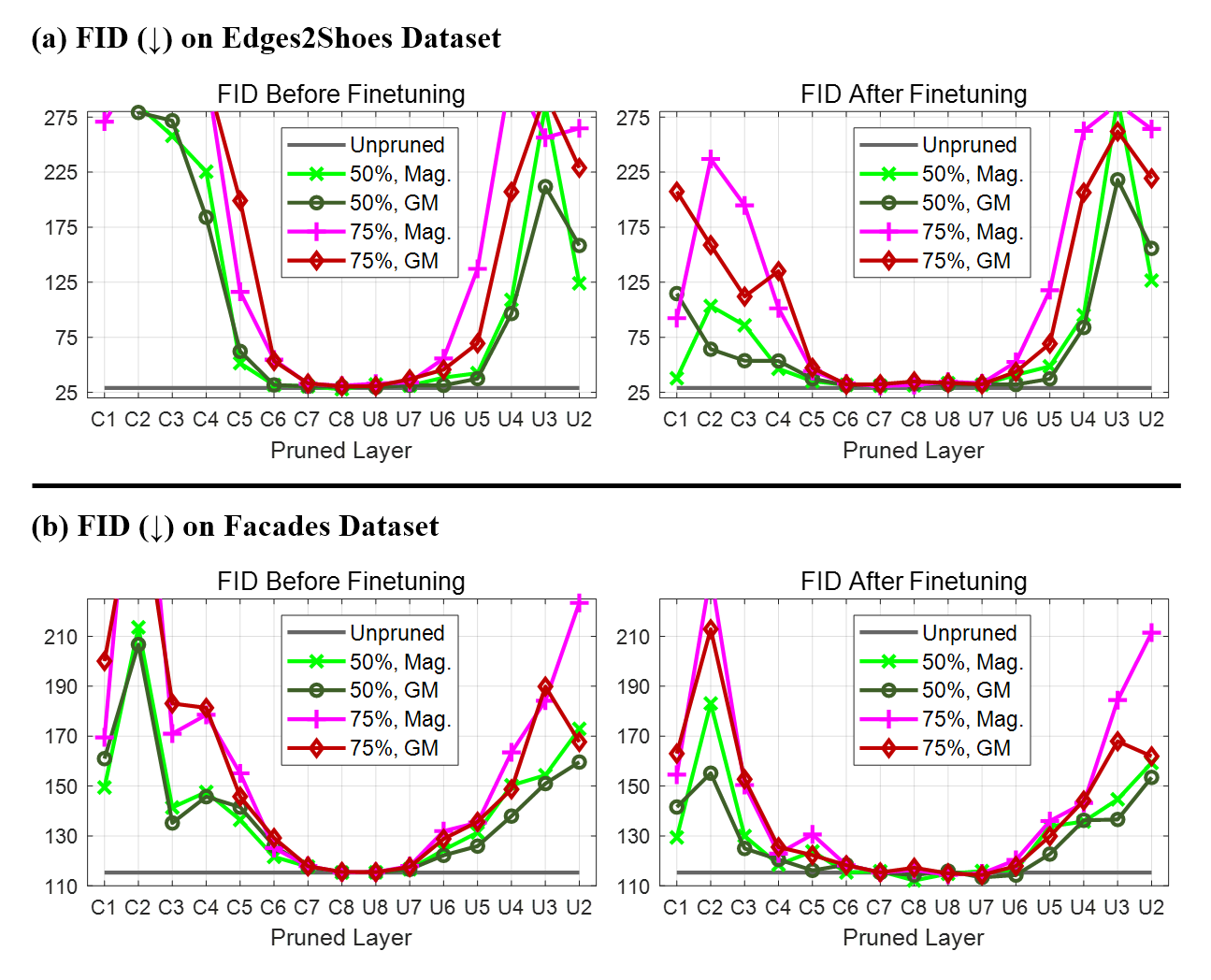}
    \caption{Comparison of different pruning criteria. `Mag.' denotes the weight magnitude criterion, and `GM' denotes the geometric median criterion. The pruning ratio is \{50\%, 75\%\}. Lower values are better.}
    \label{fig_gmComp}
\end{center}
\end{figure} 


\section{Experimental Setup} \label{sec:exp}

\textbf{Architecture.} \Cref{fig_network} depicts the U-Net generators of Pix2Pix \cite{pix2pix2017} and Wav2Lip \cite{prajwal2020lip}. For Wav2Lip, we build an efficient yet effective generator by halving the number of filters (i.e., ($\mathrm{nVF}$, $\mathrm{nAF}$, $\mathrm{nDF}$) changes from (16, 32, 32) to (8, 16, 16)) and removing all the residual blocks. See \Cref{appendix_sect_arch} for the details.

\textbf{Baseline.} We compare our method with global pruning approaches to validate the importance of properly considering where to prune in the U-Net. \textit{Uniform} pruning removes small-magnitude filters of every layer with a given pruning ratio. \textit{Uniform+} pruning is almost identical to Uniform, except that the first encoder layer and the last decoder layer are not pruned. This is similar to the heuristic suggested in \citet{gale2019state} for classification models. Moreover, we modify Layer-Adaptive Magnitude-based Pruning (LAMP) \cite{lee2020layer}, which was originally unstructured pruning, to perform structured pruning and call it \textit{Structured LAMP}. 

\textbf{Dataset.} For Pix2Pix, we consider the Edges2Shoes \cite{data_e2s} and Facades \cite{data_fcd} datasets that contain 50\textit{K} and 0.5\textit{K} images, respectively. We follow the training and test data splits of \citet{pix2pix2017}. For Wav2Lip, we use the LRS3 dataset \cite{data_lrs3} that contains 32\textit{K} utterances from 4\textit{K} speakers of TED videos. We train and evaluate the models with the original data splits of LRS3.

\textbf{Evaluation Metric.} We adopt Frechet Inception Distance (FID) \cite{heusel2017gans} to assess the generation quality. Moreover, we consider recent metrics to separately evaluate the fidelity and diversity of generated images \cite{kynkaanniemi2019improved, naeem2020reliable}. For Wav2Lip, we additionally compute Lip Sync Error - Distance (LSE-D) and Confidence (LSE-C) \cite{chung2016out} to assess the quality of lip synchronization between speech and generated face samples. We also report the number of parameters and multiply-accumulate operations (MACs)\footnote{Although MACs have been widely used to measure the computational cost, they may lead to an inaccurate energy consumption estimation \cite{8114708}.}.

\textbf{Implementation Details.} We implement the pruning methods with NetsPresso Model Compressor\footnote{https://www.netspresso.ai/model-compressor} and adopt the publicly available codes of Pix2Pix\footnote{https://github.com/junyanz/pytorch-CycleGAN-and-pix2pix} and Wav2Lip\footnote{https://github.com/Rudrabha/Wav2Lip}. The optimization settings closely follow the original settings of \citet{pix2pix2017} and \citet{prajwal2020lip} except empirically determining the batch size, the number of training epochs, and the type of GAN loss (see \Cref{appendix_sect_impl} for the details). We use a single NVIDIA GeForce RTX 3090 GPU for the training process.

\section{Results and Analyses} \label{sec:result}

\subsection{Pruning Pix2Pix Models} 

\subsubsection{Layer-wise Sensitivity Analysis}

\Cref{fig_per_layer_sensitivity} shows the sensitivity analysis. Pruning each of the five innermost layers (i.e., C6, C7, C8, U8, U7) leads to a similar FID score with the original model even at the pruning ratio of 75\%, indicating that these layers contain many unnecessary filters and are insensitive to pruning. In contrast, pruning outer layers causes substantial performance drops, confirming that outer layers and their skip connections are essential for the generation tasks. In other words, because of the same underlying structure shared between the input and output, delivering low-level information via outer skip connections is important. This U-shaped trend also appears in the case of a smaller generator capacity.

\subsubsection{Filter-level and Layer-level Pruning}

\Cref{table:e2s_result,table:facades_result} report the quantitative results. For the Edges2Shoes dataset, we were unable to achieve compelling results with the global pruning baselines. Eliminating only 1\% or 2\% of the filters from every layer with the uniform methods degrades the FID scores. The LAMP method removes many filters in the U4 layer, which is identified as a pruning-sensitive layer in our layer-wise analysis.

By contrast, filter-level and layer-level pruning of the innermost layers do not exhibit noticeable performance drops while effectively reducing the number of parameters and MACs. These results suggest that properly determining where to prune is important. Interestingly, some pruned networks yield better FID scores than the original networks, which may be connected to the ability of pruning for mitigating overfitting \cite{hanson1988comparing} and ablating defective weights \cite{Tousi_2021_CVPR}. Moreover, as shown in \Cref{fig_visualEx_pix2pix}, our pruned models perform comparatively to the original models in terms of visual fidelity.

\subsection{Pruning Wav2Lip Models} 

From the compressed generator, we prune the filters of the three innermost layers of the vision encoder, audio encoder, and decoder in \Cref{fig_network}(b). The number of filters in these layers changes from 256 or 192 to 128 after pruning. 

\Cref{table:wav2lip_result} shows the quantitative results on the LRS3 dataset. Although the compressed generator was already much more efficient than the original large model, we can further compress it without significant performance degradation. Moreover, our method outperforms the global pruning baselines that have similar computational costs, demonstrating the effectiveness of pruning innermost filters in the U-Net generator. \Cref{fig_visualEx_wav2lip} depicts some visual examples. Our method can generate lip-synced face images without loss of visual quality under a reduced computational budget.

\subsection{Additional Evaluation Metrics} \label{sect_evalMetr}

In addition to FID, we further adopt recent metrics \cite{kynkaanniemi2019improved, naeem2020reliable} to study how pruning of each layer affects the fidelity and diversity aspects of generated images. \Cref{fig_eval_metrics} shows the results on the Edges2shoes dataset with Pix2Pix. Similarly to \Cref{fig_per_layer_sensitivity}, filter pruning of the innermost layers better preserves the examined aspects than that of the outer layers. At the severe pruning ratio of 75\% before finetuning, pruning innermost filters retains the fidelity but slightly degrades the diversity, which is similarly reported in \citet{mordido2021evaluating}.

\subsection{Comparison of Pruning Criteria} \label{sect_gm}

Beyond the importance-based criterion using weight magnitude, we investigate the redundancy-based criterion using geometric median \cite{he2019filter}. \Cref{fig_gmComp} compares the two criteria in the sensitivity analysis using the large generator of Pix2Pix. A consistent trend is observed regardless of the pruning criteria: the innermost layers in the U-Net contain a large number of unnecessary filters and are highly prunable. From this observation, we can understand the network behavior that mainly utilizes the outer layers and their skip connections for fulfilling the generation task.

\section{Conclusion} \label{sec:conclusion}

We demonstrate high prunability of the inner layers in U-Net generators and present filter- and layer-level structured pruning methods that exploit the layer characteristics. We believe that the insights from our study can help understanding and improving the generator architectures for GANs. Combining our approach with knowledge distillation to boost the performance and with quantization for further compression would be a promising future direction.


\bibliography{haet22_nota}

\begin{thebibliography}{48}
\providecommand{\natexlab}[1]{#1}
\providecommand{\url}[1]{\texttt{#1}}
\expandafter\ifx\csname urlstyle\endcsname\relax
  \providecommand{\doi}[1]{doi: #1}\else
  \providecommand{\doi}{doi: \begingroup \urlstyle{rm}\Url}\fi

\bibitem[Afouras et~al.(2018)Afouras, Chung, and Zisserman]{data_lrs3}
Afouras, T., Chung, J.~S., and Zisserman, A.
\newblock Lrs3-ted: a large-scale dataset for visual speech recognition.
\newblock \emph{arXiv preprint arXiv:1809.00496}, 2018.

\bibitem[Andreev et~al.(2021)Andreev, Fritzler, and
  Vetrov]{andreev2021quantization}
Andreev, P., Fritzler, A., and Vetrov, D.
\newblock Quantization of generative adversarial networks for efficient
  inference: a methodological study.
\newblock \emph{arXiv preprint arXiv:2108.13996}, 2021.

\bibitem[Chen et~al.(2021{\natexlab{a}})Chen, Cheng, Gan, Liu, and
  Wang]{chen2021data}
Chen, T., Cheng, Y., Gan, Z., Liu, J., and Wang, Z.
\newblock Data-efficient gan training beyond (just) augmentations: A lottery
  ticket perspective.
\newblock In \emph{NeurIPS}, 2021{\natexlab{a}}.

\bibitem[Chen et~al.(2021{\natexlab{b}})Chen, Zhang, Sui, and
  Chen]{chen2021gans}
Chen, X., Zhang, Z., Sui, Y., and Chen, T.
\newblock Gans can play lottery tickets too.
\newblock In \emph{ICLR}, 2021{\natexlab{b}}.

\bibitem[Chung \& Zisserman(2016)Chung and Zisserman]{chung2016out}
Chung, J.~S. and Zisserman, A.
\newblock Out of time: automated lip sync in the wild.
\newblock In \emph{ACCV}, 2016.

\bibitem[Frankle \& Carbin(2019)Frankle and Carbin]{frankle2018lottery}
Frankle, J. and Carbin, M.
\newblock The lottery ticket hypothesis: Finding sparse, trainable neural
  networks.
\newblock In \emph{ICLR}, 2019.

\bibitem[Fu et~al.(2020)Fu, Chen, Wang, Li, Lin, and Wang]{fu2020autogan}
Fu, Y., Chen, W., Wang, H., Li, H., Lin, Y., and Wang, Z.
\newblock Autogan-distiller: Searching to compress generative adversarial
  networks.
\newblock In \emph{ICML}, 2020.

\bibitem[Gale et~al.(2019)Gale, Elsen, and Hooker]{gale2019state}
Gale, T., Elsen, E., and Hooker, S.
\newblock The state of sparsity in deep neural networks.
\newblock In \emph{ICML Workshop}, 2019.

\bibitem[Gao et~al.(2020)Gao, Chen, Liu, Tan, and Yan]{gao2020adversarialnas}
Gao, C., Chen, Y., Liu, S., Tan, Z., and Yan, S.
\newblock Adversarialnas: Adversarial neural architecture search for gans.
\newblock In \emph{CVPR}, 2020.

\bibitem[Ghosh et~al.(2019)Ghosh, Srinivasa, Amon, Hutter, and
  Kaup]{ghosh2019deep}
Ghosh, S., Srinivasa, S.~K., Amon, P., Hutter, A., and Kaup, A.
\newblock Deep network pruning for object detection.
\newblock In \emph{ICIP}, 2019.

\bibitem[Gong et~al.(2019)Gong, Chang, Jiang, and Wang]{gong2019autogan}
Gong, X., Chang, S., Jiang, Y., and Wang, Z.
\newblock Autogan: Neural architecture search for generative adversarial
  networks.
\newblock In \emph{ICCV}, 2019.

\bibitem[Goodfellow et~al.(2014)Goodfellow, Pouget-Abadie, Mirza, Xu,
  Warde-Farley, Ozair, Courville, and Bengio]{goodfellow2014generative}
Goodfellow, I., Pouget-Abadie, J., Mirza, M., Xu, B., Warde-Farley, D., Ozair,
  S., Courville, A., and Bengio, Y.
\newblock Generative adversarial nets.
\newblock In \emph{NeurIPS}, 2014.

\bibitem[Hanson \& Pratt(1988)Hanson and Pratt]{hanson1988comparing}
Hanson, S. and Pratt, L.
\newblock Comparing biases for minimal network construction with
  back-propagation.
\newblock In \emph{NeurIPS}, 1988.

\bibitem[He et~al.(2019)He, Liu, Wang, Hu, and Yang]{he2019filter}
He, Y., Liu, P., Wang, Z., Hu, Z., and Yang, Y.
\newblock Filter pruning via geometric median for deep convolutional neural
  networks acceleration.
\newblock In \emph{CVPR}, 2019.

\bibitem[Heusel et~al.(2017)Heusel, Ramsauer, Unterthiner, Nessler, and
  Hochreiter]{heusel2017gans}
Heusel, M., Ramsauer, H., Unterthiner, T., Nessler, B., and Hochreiter, S.
\newblock Gans trained by a two time-scale update rule converge to a local nash
  equilibrium.
\newblock In \emph{NeurIPS}, 2017.

\bibitem[Hinton et~al.(2014)Hinton, Vinyals, and Dean]{hinton2015distilling}
Hinton, G., Vinyals, O., and Dean, J.
\newblock Distilling the knowledge in a neural network.
\newblock In \emph{NeurIPS Workshop}, 2014.

\bibitem[Hou et~al.(2021)Hou, Yuan, Huang, Shen, Cheng, and
  Wang]{hou2020slimmable}
Hou, L., Yuan, Z., Huang, L., Shen, H., Cheng, X., and Wang, C.
\newblock Slimmable generative adversarial networks.
\newblock In \emph{AAAI}, 2021.

\bibitem[Isola et~al.(2017)Isola, Zhu, Zhou, and Efros]{pix2pix2017}
Isola, P., Zhu, J.-Y., Zhou, T., and Efros, A.~A.
\newblock Image-to-image translation with conditional adversarial networks.
\newblock In \emph{CVPR}, 2017.

\bibitem[Jin et~al.(2021)Jin, Ren, Woodford, Wang, Yuan, Wang, and
  Tulyakov]{jin2021teachers}
Jin, Q., Ren, J., Woodford, O.~J., Wang, J., Yuan, G., Wang, Y., and Tulyakov,
  S.
\newblock Teachers do more than teach: Compressing image-to-image models.
\newblock In \emph{CVPR}, 2021.

\bibitem[Kynk{\"a}{\"a}nniemi et~al.(2019)Kynk{\"a}{\"a}nniemi, Karras, Laine,
  Lehtinen, and Aila]{kynkaanniemi2019improved}
Kynk{\"a}{\"a}nniemi, T., Karras, T., Laine, S., Lehtinen, J., and Aila, T.
\newblock Improved precision and recall metric for assessing generative models.
\newblock In \emph{NeurIPS}, 2019.

\bibitem[LeCun et~al.(1989)LeCun, Denker, and Solla]{lecun1989optimal}
LeCun, Y., Denker, J., and Solla, S.
\newblock Optimal brain damage.
\newblock In \emph{NeurIPS}, 1989.

\bibitem[Lee et~al.(2021)Lee, Park, Mo, Ahn, and Shin]{lee2020layer}
Lee, J., Park, S., Mo, S., Ahn, S., and Shin, J.
\newblock Layer-adaptive sparsity for the magnitude-based pruning.
\newblock In \emph{ICLR}, 2021.

\bibitem[Li et~al.(2017)Li, Kadav, Durdanovic, Samet, and Graf]{li2016pruning}
Li, H., Kadav, A., Durdanovic, I., Samet, H., and Graf, H.~P.
\newblock Pruning filters for efficient convnets.
\newblock In \emph{ICLR}, 2017.

\bibitem[Li et~al.(2020)Li, Lin, Ding, Liu, Zhu, and Han]{li2020gan}
Li, M., Lin, J., Ding, Y., Liu, Z., Zhu, J.-Y., and Han, S.
\newblock Gan compression: Efficient architectures for interactive conditional
  gans.
\newblock In \emph{CVPR}, 2020.

\bibitem[Li et~al.(2021)Li, Wu, Xiao, Chao, Mao, and Ji]{li2021revisiting}
Li, S., Wu, J., Xiao, X., Chao, F., Mao, X., and Ji, R.
\newblock Revisiting discriminator in gan compression: A
  generator-discriminator cooperative compression scheme.
\newblock In \emph{NeurIPS}, 2021.

\bibitem[Li et~al.(2022)Li, Lin, Wang, Fei, Shao, and Ji]{li2022learning}
Li, S., Lin, M., Wang, Y., Fei, C., Shao, L., and Ji, R.
\newblock Learning efficient gans for image translation via differentiable
  masks and co-attention distillation.
\newblock \emph{IEEE Trans. Multimedia}, 2022.

\bibitem[Lin et~al.(2019)Lin, Ji, Li, Deng, and Li]{lin2019toward}
Lin, S., Ji, R., Li, Y., Deng, C., and Li, X.
\newblock Toward compact convnets via structure-sparsity regularized filter
  pruning.
\newblock \emph{IEEE Trans. Neural Netw. Learn. Syst.}, 31\penalty0
  (2):\penalty0 574--588, 2019.

\bibitem[Liu et~al.(2021)Liu, Shu, Li, Lin, Perazzi, and Kung]{liu2021content}
Liu, Y., Shu, Z., Li, Y., Lin, Z., Perazzi, F., and Kung, S.-Y.
\newblock Content-aware gan compression.
\newblock In \emph{CVPR}, 2021.

\bibitem[Mordido et~al.(2021)Mordido, Yang, and Meinel]{mordido2021evaluating}
Mordido, G., Yang, H., and Meinel, C.
\newblock Evaluating post-training compression in gans using locality-sensitive
  hashing.
\newblock \emph{arXiv preprint arXiv:2103.11912}, 2021.

\bibitem[Naeem et~al.(2020)Naeem, Oh, Uh, Choi, and Yoo]{naeem2020reliable}
Naeem, M.~F., Oh, S.~J., Uh, Y., Choi, Y., and Yoo, J.
\newblock Reliable fidelity and diversity metrics for generative models.
\newblock In \emph{ICML}, 2020.

\bibitem[Ning et~al.(2020)Ning, Zhao, Li, Lei, Wang, and Yang]{ning2020dsa}
Ning, X., Zhao, T., Li, W., Lei, P., Wang, Y., and Yang, H.
\newblock Dsa: More efficient budgeted pruning via differentiable sparsity
  allocation.
\newblock In \emph{ECCV}, 2020.

\bibitem[Prajwal et~al.(2020)Prajwal, Mukhopadhyay, Namboodiri, and
  Jawahar]{prajwal2020lip}
Prajwal, K., Mukhopadhyay, R., Namboodiri, V.~P., and Jawahar, C.
\newblock A lip sync expert is all you need for speech to lip generation in the
  wild.
\newblock In \emph{ACM MM}, 2020.

\bibitem[Romero et~al.(2015)Romero, Ballas, Kahou, Chassang, Gatta, and
  Bengio]{romero2014fitnets}
Romero, A., Ballas, N., Kahou, S.~E., Chassang, A., Gatta, C., and Bengio, Y.
\newblock Fitnets: Hints for thin deep nets.
\newblock In \emph{ICLR}, 2015.

\bibitem[Ronneberger et~al.(2015)Ronneberger, Fischer, and
  Brox]{ronneberger2015u}
Ronneberger, O., Fischer, P., and Brox, T.
\newblock U-net: Convolutional networks for biomedical image segmentation.
\newblock In \emph{MICCAI}, 2015.

\bibitem[Shih et~al.(2019)Shih, Chiu, and Pu]{shih2019real}
Shih, K.-H., Chiu, C.-T., and Pu, Y.-Y.
\newblock Real-time object detection via pruning and a concatenated
  multi-feature assisted region proposal network.
\newblock In \emph{ICASSP}, 2019.

\bibitem[Shu et~al.(2019)Shu, Wang, Jia, Han, Chen, Xu, Tian, and
  Xu]{shu2019co}
Shu, H., Wang, Y., Jia, X., Han, K., Chen, H., Xu, C., Tian, Q., and Xu, C.
\newblock Co-evolutionary compression for unpaired image translation.
\newblock In \emph{ICCV}, 2019.

\bibitem[Sze et~al.(2017)Sze, Chen, Yang, and Emer]{8114708}
Sze, V., Chen, Y.-H., Yang, T.-J., and Emer, J.~S.
\newblock Efficient processing of deep neural networks: A tutorial and survey.
\newblock \emph{Proc. IEEE}, 105\penalty0 (12):\penalty0 2295--2329, 2017.

\bibitem[Tousi et~al.(2021)Tousi, Jeong, Han, Choi, and Choi]{Tousi_2021_CVPR}
Tousi, A., Jeong, H., Han, J., Choi, H., and Choi, J.
\newblock Automatic correction of internal units in generative neural networks.
\newblock In \emph{CVPR}, 2021.

\bibitem[Tyle{\v{c}}ek \& {\v{S}}{\'a}ra(2013)Tyle{\v{c}}ek and
  {\v{S}}{\'a}ra]{data_fcd}
Tyle{\v{c}}ek, R. and {\v{S}}{\'a}ra, R.
\newblock Spatial pattern templates for recognition of objects with regular
  structure.
\newblock In \emph{GCPR}, 2013.

\bibitem[Wan et~al.(2020)Wan, Shen, Liu, Zhu, Huang, Yu, Shen, and
  Shao]{wan2020deep}
Wan, D., Shen, F., Liu, L., Zhu, F., Huang, L., Yu, M., Shen, H.~T., and Shao,
  L.
\newblock Deep quantization generative networks.
\newblock \emph{Pattern Recognition}, 105:\penalty0 107338, 2020.

\bibitem[Wang et~al.(2020)Wang, Gui, Yang, Liu, and Wang]{wang2020gan}
Wang, H., Gui, S., Yang, H., Liu, J., and Wang, Z.
\newblock Gan slimming: All-in-one gan compression by a unified optimization
  framework.
\newblock In \emph{ECCV}, 2020.

\bibitem[Wang et~al.(2019)Wang, Wang, Ji, Xie, Song, Liu, Lyu, and
  Xie]{wang2019qgan}
Wang, P., Wang, D., Ji, Y., Xie, X., Song, H., Liu, X., Lyu, Y., and Xie, Y.
\newblock Qgan: Quantized generative adversarial networks.
\newblock \emph{arXiv preprint arXiv:1901.08263}, 2019.

\bibitem[Wen et~al.(2016)Wen, Wu, Wang, Chen, and Li]{wen2016learning}
Wen, W., Wu, C., Wang, Y., Chen, Y., and Li, H.
\newblock Learning structured sparsity in deep neural networks.
\newblock In \emph{NeurIPS}, 2016.

\bibitem[Xie et~al.(2020)Xie, Zhu, Zhao, Tao, Liu, and
  Tao]{xie2020localization}
Xie, Z., Zhu, L., Zhao, L., Tao, B., Liu, L., and Tao, W.
\newblock Localization-aware channel pruning for object detection.
\newblock \emph{Neurocomputing}, 403:\penalty0 400--408, 2020.

\bibitem[Yang et~al.(2020)Yang, Zhang, Guo, Liu, Zuo, and Luo]{yang2020towards}
Yang, H., Zhang, R., Guo, X., Liu, W., Zuo, W., and Luo, P.
\newblock Towards photo-realistic virtual try-on by adaptively
  generating-preserving image content.
\newblock In \emph{CVPR}, 2020.

\bibitem[Yeom et~al.(2021)Yeom, Seegerer, Lapuschkin, Binder, Wiedemann,
  M{\"u}ller, and Samek]{yeom2021pruning}
Yeom, S.-K., Seegerer, P., Lapuschkin, S., Binder, A., Wiedemann, S.,
  M{\"u}ller, K.-R., and Samek, W.
\newblock Pruning by explaining: A novel criterion for deep neural network
  pruning.
\newblock \emph{Pattern Recognition}, 115:\penalty0 107899, 2021.

\bibitem[Yu \& Grauman(2014)Yu and Grauman]{data_e2s}
Yu, A. and Grauman, K.
\newblock Fine-grained visual comparisons with local learning.
\newblock In \emph{CVPR}, 2014.

\bibitem[Yu \& Pool(2020)Yu and Pool]{yu2020self}
Yu, C. and Pool, J.
\newblock Self-supervised generative adversarial compression.
\newblock In \emph{NeurIPS}, 2020.

\end{thebibliography}
\bibliographystyle{icml2022}

\newpage
\appendix

\section*{Appendix}

\section{Generator Architectures}  \label{appendix_sect_arch}

\subsection{Pix2Pix}

Pix2Pix \cite{pix2pix2017} is a popular conditional GAN for image-to-image translation tasks. \Cref{fig_network}(a) shows its U-Net generator. The encoder consists of convolutional layers that progressively downsample the input to capture the global image context, and the decoder consists of transposed convolutional layers that progressively upsample the bottleneck features to produce the output. The spatial size of the bottleneck feature maps becomes 1×1. Skip connections are used to deliver the encoded feature maps directly to the decoder, improving the generation quality. We test our method for the original large-capacity generator ($\textrm{nF}=64$) and a smaller one ($\textrm{nF}=32$).

\subsection{Wav2Lip}

Wav2Lip \cite{prajwal2020lip} is a GAN designed for speech-driven talking face generation. Unlike the single-path encoder of Pix2Pix, its generator contains a double-path encoder, as shown in \Cref{fig_network}(b). The audio encoder takes a speech segment as input, and the face encoder takes a reference frame concatenated with a pose-prior frame. Both encoders map their inputs to the 1×1 embedding vectors, which are concatenated at the bottleneck. The decoder produces a face image of the reference identity that matches the given speech with proper lip synchronization. Skip connections are added between the face encoder and the decoder.

We build an efficient yet effective generator by halving the number of filters (i.e., ($\mathrm{nVF}$, $\mathrm{nAF}$, $\mathrm{nDF}$) changes from (16, 32, 32) to (8, 16, 16)) and removing all the residual blocks from the original model. This compressed generator is trained from scratch and becomes the target of pruning.

\section{Comparison of Retraining Scenarios} \label{appendix_sect_retrain}

We compare the following settings for retraining pruned generators with adversarial learning. For brevity, $G$ and $D$ denote the generator and discriminator, respectively.

\begin{itemize}
  \item (i) Motivated by \citet{frankle2018lottery}, we re-initialize the pruned $G$ with random weights. Considering the balance between $G$ and $D$, we also re-initialize $D$ and jointly train $G$ and $D$ from scratch.
  \item (ii) We use the pruned $G$ with its pretrained weights as initialization. The pretrained $D$ is freezed (not updated) but provides learning signals for retraining $G$.
  \item (iii) This setting is the same as `(ii)' except that the pretrained $D$ is jointly trained with the pruned $G$. We use it throughout this work.
\end{itemize}

\Cref{table_retrain} summarizes the results of the three scenarios for layer-level pruning. Regardless of the type of removed innermost layers, `(ii)' performs worse than `(i)' and `(iii)', indicating that the pruned generators cannot receive meaningful gradients from the freezed discriminators.\footnote{A possible reason why `(ii)' fails is as follows: a well-pretrained $D$ cannot distinguish real and generated images and produces the output probability of 0.5 for both image types. In `(ii)', $G$ being retrained generates the images whose probability becomes assigned as 1 by $D$. This causes a distribution shift of $D$'s outputs and may not yield helpful gradients for retraining $G$.} In addition, `(ii)' does not match the concept of adversarial learning where the balance between the generator and discriminator must be well kept.

In our experiments, `(iii)' consistently outperforms `(i)' for different datasets and pruned architectures, suggesting that finetuning pretrained discriminators jointly with pruned generators is beneficial.

\begin{table}[t]
\centering
\begin{adjustbox}{max width=\columnwidth}

\begin{tabular}{cccccc}

\specialrule{.2em}{.1em}{.1em} 
\multicolumn{6}{c}{FID (↓) on Edges2Shoes Dataset}                                                                                                                                                                                                  \\ 

\hline
\noalign{\vskip\doublerulesep
         \vskip-\arrayrulewidth}
\hline

\multicolumn{1}{c|}{\multirow{2}{*}{\begin{tabular}[c]{@{}c@{}}Layer\\ Removal\end{tabular}}}     & \multicolumn{2}{c|}{\# Removed Innermost Layers}                        & \multicolumn{1}{c|}{2}     & \multicolumn{1}{c|}{4}     & 6     \\ \cline{2-6} 
\multicolumn{1}{c|}{}                                                                             & \multicolumn{2}{c|}{Bottleneck Size}                                    & \multicolumn{1}{c|}{2×2}   & \multicolumn{1}{c|}{4×4}   & 8×8   \\ 

\hline
\noalign{\vskip\doublerulesep
         \vskip-\arrayrulewidth}
\hline

\multicolumn{1}{c|}{\multirow{3}{*}{\begin{tabular}[c]{@{}c@{}}Retraining\\ Scenario\end{tabular}}} & \multicolumn{1}{c|}{(i)} & \multicolumn{1}{c|}{Rand. Init. + Train $D$}  & \multicolumn{1}{c|}{31.8}  & \multicolumn{1}{c|}{32.2}  & 63.7  \\ \cline{2-6} 
\multicolumn{1}{c|}{}                                                                             & \multicolumn{1}{c|}{(ii)}  & \multicolumn{1}{c|}{Pretrained Init. + Freeze $D$} & \multicolumn{1}{c|}{47.8}  & \multicolumn{1}{c|}{71.5}  & 108.7 \\ \cline{2-6} 
\multicolumn{1}{c|}{}                                                                             & \multicolumn{1}{c|}{(iii)}  & \multicolumn{1}{c|}{Pretrained Init. + Train $D$}  & \multicolumn{1}{c|}{28.3}  & \multicolumn{1}{c|}{29.7}  & 38.7  \\ 

\specialrule{.2em}{.1em}{.1em} 

\specialrule{.2em}{.1em}{.1em} 

\multicolumn{6}{c}{FID (↓) on Facades Dataset}                                                                                                                                                                                                      \\
\hline
\noalign{\vskip\doublerulesep
         \vskip-\arrayrulewidth}
\hline

\multicolumn{1}{c|}{\multirow{2}{*}{\begin{tabular}[c]{@{}c@{}}Layer\\ Removal\end{tabular}}}     & \multicolumn{2}{c|}{\# Removed Innermost Layers}                        & \multicolumn{1}{c|}{2}     & \multicolumn{1}{c|}{4}     & 6     \\ \cline{2-6} 
\multicolumn{1}{c|}{}                                                                             & \multicolumn{2}{c|}{Bottleneck Size}                                    & \multicolumn{1}{c|}{2×2}   & \multicolumn{1}{c|}{4×4}   & 8×8   \\ 

\hline
\noalign{\vskip\doublerulesep
         \vskip-\arrayrulewidth}
\hline

\multicolumn{1}{c|}{\multirow{3}{*}{\begin{tabular}[c]{@{}c@{}}Retraining\\ Scenario\end{tabular}}} & \multicolumn{1}{c|}{(i)} & \multicolumn{1}{c|}{Rand. Init. + Train $D$}  & \multicolumn{1}{c|}{115.2} & \multicolumn{1}{c|}{112.2} & 113.5 \\ \cline{2-6} 
\multicolumn{1}{c|}{}                                                                             & \multicolumn{1}{c|}{(ii)}  & \multicolumn{1}{c|}{Pretrained Init. + Freeze $D$} & \multicolumn{1}{c|}{158.5} & \multicolumn{1}{c|}{159.0} & 181.8 \\ \cline{2-6} 
\multicolumn{1}{c|}{}                                                                             & \multicolumn{1}{c|}{(iii)}  & \multicolumn{1}{c|}{Pretrained Init. + Train $D$}  & \multicolumn{1}{c|}{107.9} & \multicolumn{1}{c|}{108.7} & 107.8 \\ 

\specialrule{.2em}{.1em}{.1em} 

\end{tabular}

\end{adjustbox}

\caption{Comparison of different settings for retraining layer-pruned generators.}
\label{table_retrain}
\end{table}

\section{Implementation Details}  \label{appendix_sect_impl}

We use the following hyperparameters in the experiments:

\begin{itemize}[noitemsep,topsep=0pt,parsep=0pt,partopsep=0pt,font=\small]
    \item Common setup for Pix2Pix models
        \begin{itemize}[noitemsep,topsep=0pt,parsep=0pt,partopsep=0pt]
        \item Initial learning rate (LR): 0.0002
        \item GAN loss: Hinge
        \item Optimizer: Adam with ($\beta1$, $\beta2$)=(0.5, 0.999)
    \end{itemize}
\end{itemize}

\begin{itemize}[noitemsep,topsep=0pt,parsep=0pt,partopsep=0pt,font=\small]
    \item Pix2Pix on the Edge2Shoes dataset
        \begin{itemize}[noitemsep,topsep=0pt,parsep=0pt,partopsep=0pt]
        \item Batch size: 32 
        \item \# (Pretraining, Retraining) epochs: (200, 15)
        \item LR schedule: Linear decay to zero after the (100, 10)-th epoch for (pretraining, retraining)
    \end{itemize}
\end{itemize}

\begin{itemize}[noitemsep,topsep=0pt,parsep=0pt,partopsep=0pt,font=\small]
    \item Pix2Pix on the Facades dataset
        \begin{itemize}[noitemsep,topsep=0pt,parsep=0pt,partopsep=0pt]
        \item Batch size: 1
        \item \# (Pretraining, Retraining) epochs: (300, 20)
        \item LR schedule: Linear decay to zero after the (200, 15)-th epoch for (pretraining, retraining)
    \end{itemize}
\end{itemize}

\begin{itemize}[noitemsep,topsep=0pt,parsep=0pt,partopsep=0pt,font=\small]
    \item Wav2Lip on the LRS3 dataset
        \begin{itemize}[noitemsep,topsep=0pt,parsep=0pt,partopsep=0pt]
        \item Initial LR: 0.0001
        \item GAN loss: Standard
        \item Optimizer: Adam with ($\beta1$, $\beta2$)=(0.5, 0.999)
        \item Batch size: 16
        \item \# (Pretraining, Retraining) epochs: (270, 50)
        \item LR schedule: N/A
    \end{itemize}
\end{itemize}


\end{document}